\documentclass[letterpaper, 10 pt, conference]{ieeeconf}  % Comment this 
\IEEEoverridecommandlockouts                              % This command 
\usepackage{amsmath,amssymb,amsfonts}
\usepackage{mathtools}      % amsmath 增强（如 \coloneqq、

\usepackage{graphicx}       % 现代插图接口（替代 epsfig）
\usepackage{booktabs}       % 表格横线
\usepackage{tabularx}       % 自适应列宽表格
\usepackage{multirow}       % 表格行合并

\usepackage{cite}
\usepackage{multirow}  % 让 Modalities 上下居中

\usepackage{algorithm}   % 提供浮动体 algorithm
\usepackage{algpseudocode} % 提供 algorithmic 和 \State/\For
\floatname{algorithm}{Algorithm} % IEEE 喜欢英文标题

\usepackage{hyperref}

\usepackage{bm}
\usepackage{times}
\usepackage{epsfig}
\usepackage{amssymb}
\usepackage{algpseudocode}
\usepackage{algorithm}
\usepackage{amsmath}
\usepackage{subfigure}
\usepackage{multirow}
\usepackage{multicol}
\usepackage{caption}
\usepackage{gensymb}
\usepackage{adjustbox}
\usepackage{mathtools}
\usepackage{booktabs}
\usepackage{pifont}
\usepackage{xspace}
\usepackage{tabularx} 
\usepackage{multirow}
\usepackage{enumerate}
\usepackage{xcolor}
\usepackage{url}
\usepackage[normalem]{ulem}
\usepackage{hyperref}

\usepackage{booktabs}
\usepackage{siunitx}
\usepackage{threeparttable}
\usepackage{makecell} % 更灵活的换行表头，可选
\usepackage{cite} % <-- 在您的导言区加上这一行！

\usepackage{hyperref}

\usepackage{booktabs}
\title{\LARGE \bf
CDFlow: Generative Gradient Flows for Configuration Space Distance Fields via Neural ODEs}

\author{
Mengzhu Li$^{1}$, Yunyu Zhou$^{2}$, He Ying$^{3}$, and F. Richard Yu$^{4}$%
\thanks{$^{1}$M. Li(corresponding author) is with the Guangdong Laboratory of Artificial Intelligence and Digital Economy (SZ), Shenzhen University, P.R. China.
{\tt\small limengzhu@gml.ac.cn.}}%
\thanks{$^{2}$Y. Zhou is with the College of Design and Engineering, National University of Singapore (NUS), Singapore.
{\tt\small yunyuzhou@u.nus.edu.}}%
\thanks{$^{3}$Y. He is with the College of Computer Science and Software Engineering, Shenzhen University, P.R. China.
{\tt\small heying@szu.edu.cn.}}%
\thanks{$^{4}$F. Richard Yu  is with the School of Information Technology, Carleton University, Canada.
{\tt\small richard.yu@ieee.org.}}%
}

\begin{document}

\maketitle
\thispagestyle{empty}
\pagestyle{empty}

%%%%%%%%%%%%%%%%%%%%%%%%%%%%%%%%%%%%%%%%%%%%%%%%%%%%%%%%%%%%%%%%%%%%%%%%%%%%%%%%

\begin{abstract}
Signed Distance Fields (SDFs) are a fundamental representation in robot motion planning. Their configuration-space counterpart, the Configuration Space Distance Field (CDF), directly encodes distances in joint space, offering a unified representation for optimization and control. However, existing CDF formulations face two major challenges in high-degree-of-freedom (DoF) robots: (1) they effectively return only a single nearest collision configuration, neglecting the multi-modal nature of \textbf{minimal-distance collision configurations} and leading to gradient ambiguity; and (2) they rely on sparse sampling of the collision boundary, which often fails to identify the true closest configurations, producing oversmoothed approximations and geometric distortion in high-dimensional spaces.

We propose \textbf{CDFlow}, a novel framework that addresses these limitations by learning a continuous flow in configuration space via Neural Ordinary Differential Equations (Neural ODEs). We redefine the problem from finding a single nearest point to modeling the \textbf{distribution of minimal-distance collision configurations}. We also introduce an adaptive refinement sampling strategy to generate high-fidelity training data for this distribution. The resulting Neural ODE implicitly models this multi-modal distribution and produces a smooth, consistent \textbf{gradient field}---derived as the expected direction towards the distribution---that mitigates gradient ambiguity and preserves sharp geometric features.

Extensive experiments on high-DoF motion planning tasks demonstrate that CDFlow significantly improves planning efficiency, trajectory quality, and robustness compared to existing CDF-based methods, enabling more robust and efficient planning for collision-aware robots in complex environments.
\end{abstract}

% 符号距离场 (SDFs) 是机器人运动规划中的一种基础表示方法。其在配置空间中的对应物——配置空间距离场 (CDF)，通过直接编码关节空间中的距离，为优化与控制提供了一种统一的表示。然而，现有的CDF公式在应用于高自由度 (DoF) 机器人时面临两大挑战：(1) 它们实际上只返回单个最近的碰撞位形，忽略了最小距离碰撞位形的多模态特性，从而导致梯度模糊性；(2) 它们依赖于对碰撞边界的稀疏采样，这常常无法识别出真正的最近位形，进而在高维空间中产生过度平滑的近似和几何失真。
% 我们提出了 CDFlow，一个通过神经微分方程 (Neural ODEs) 学习配置空间中连续流场的新颖框架，以解决上述局限。我们将问题从“寻找单个最近点”重新定义为“对最小距离碰撞位形的分布进行建模”。同时，我们引入了一种自适应精化采样策略，为该分布的训练生成高保真度样本。由此产生的Neural ODE能够隐式地建模这种多模态分布，并生成一个平滑且一致的梯度场——该梯度场通过计算朝向该分布的期望方向而导出——从而有效缓解梯度模糊性并保留锐利的几何特征。
% 在多种高自由度运动规划任务上进行的大量实验表明，与现有的基于CDF的方法相比，CDFFlow在规划效率、轨迹质量和鲁棒性方面均有显著提升，为复杂环境中具备碰撞感知能力的机器人实现了更鲁棒、更高效的规划。

\section{Introduction}

As robots are increasingly deployed in diverse fields, the ability of high-degree-of-freedom (DoF) robots to perform safe and efficient motion planning in complex, unstructured environments has become a central bottleneck to their autonomy~\cite{lavalle2006planning}. At the core of these planning tasks lies a high-dimensional geometric challenge: generating smooth, collision-free trajectories that satisfy task and environmental constraints~\cite{lynch2017modern}. A key to solving this challenge is an accurate and efficient representation of the geometric relationship between the robot and its environment. In recent years, Signed Distance Fields (SDFs) have emerged as a powerful tool for geometric representation, offering continuous distance metrics and analytical gradients~\cite{park2019deepsdf}. They have enabled significant progress in path planning~\cite{ratliff2009chomp}, motion optimization~\cite{schulman2014motion}, and robot learning~\cite{driess2022learning}.

While SDFs are conventionally defined in task space, robot planning and control must ultimately be executed in the configuration space (C-space)~\cite{lynch2017modern}. For a given robotic system and a static environment, which we collectively term the scene, the set of all colliding configurations is denoted as $\mathcal{Q}_c(\text{scene})$. The Configuration Space Distance Field (CDF)~\cite{li2024configuration} was introduced to directly encode this collision geometry in C-space. It provides structured gradients that can be used for optimization and control without repeated transformations, marking an important step toward bridging task-space geometry and configuration-space planning.

\begin{figure*}[t]
\centering
% 2 : 2 : 1
\begin{minipage}{0.4\linewidth}   % (a)
  \includegraphics[width=0.48\linewidth]{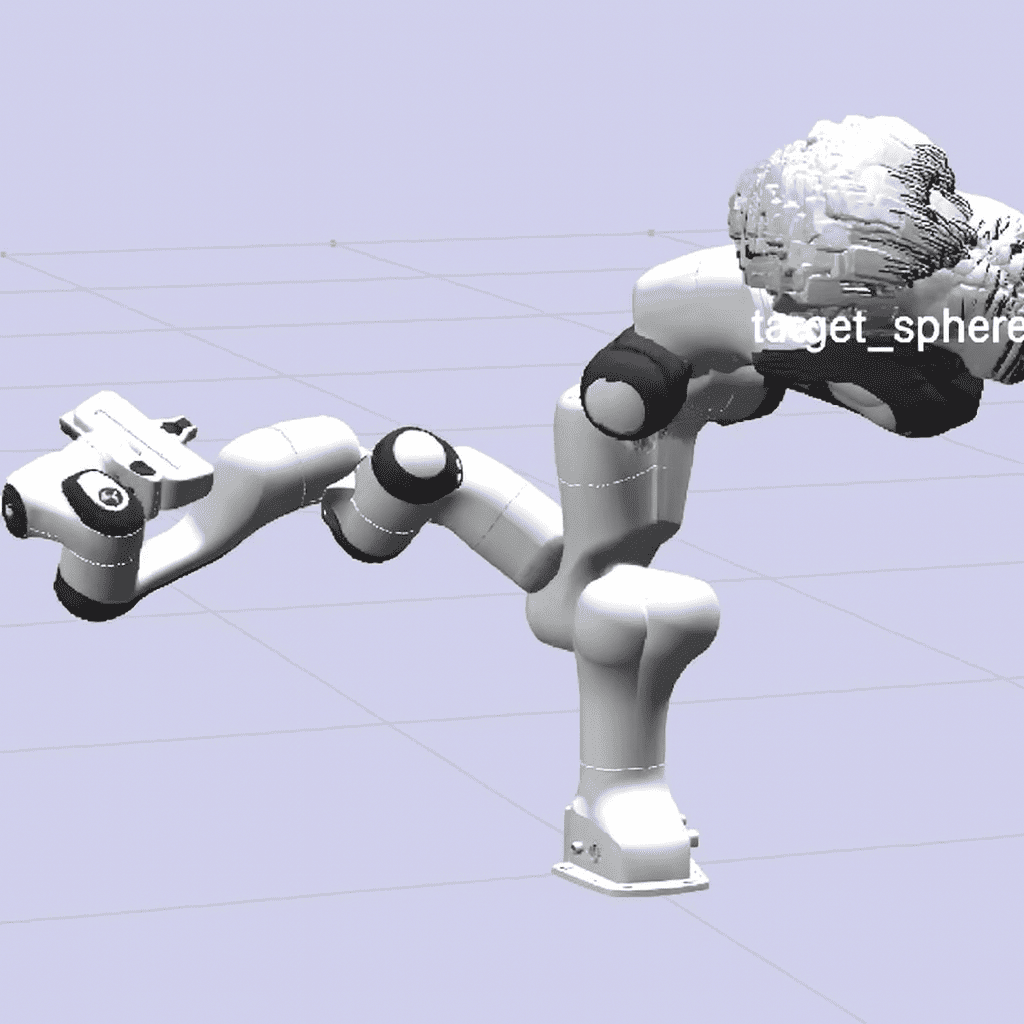}%
  \hfill
  \includegraphics[width=0.48\linewidth]{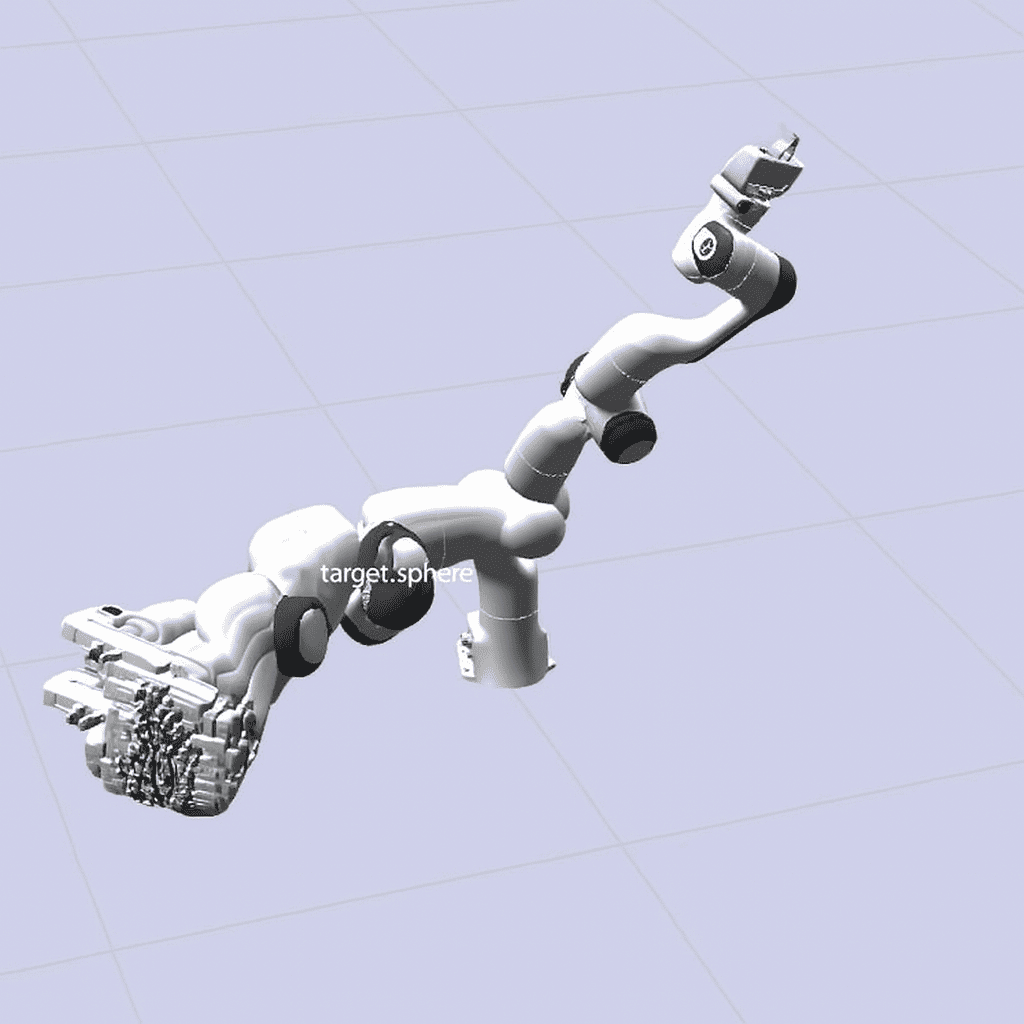}\\[2mm]
  \centering (a)
\end{minipage}\hfill
\begin{minipage}{0.4\linewidth}   % (b)
  \includegraphics[width=\linewidth]{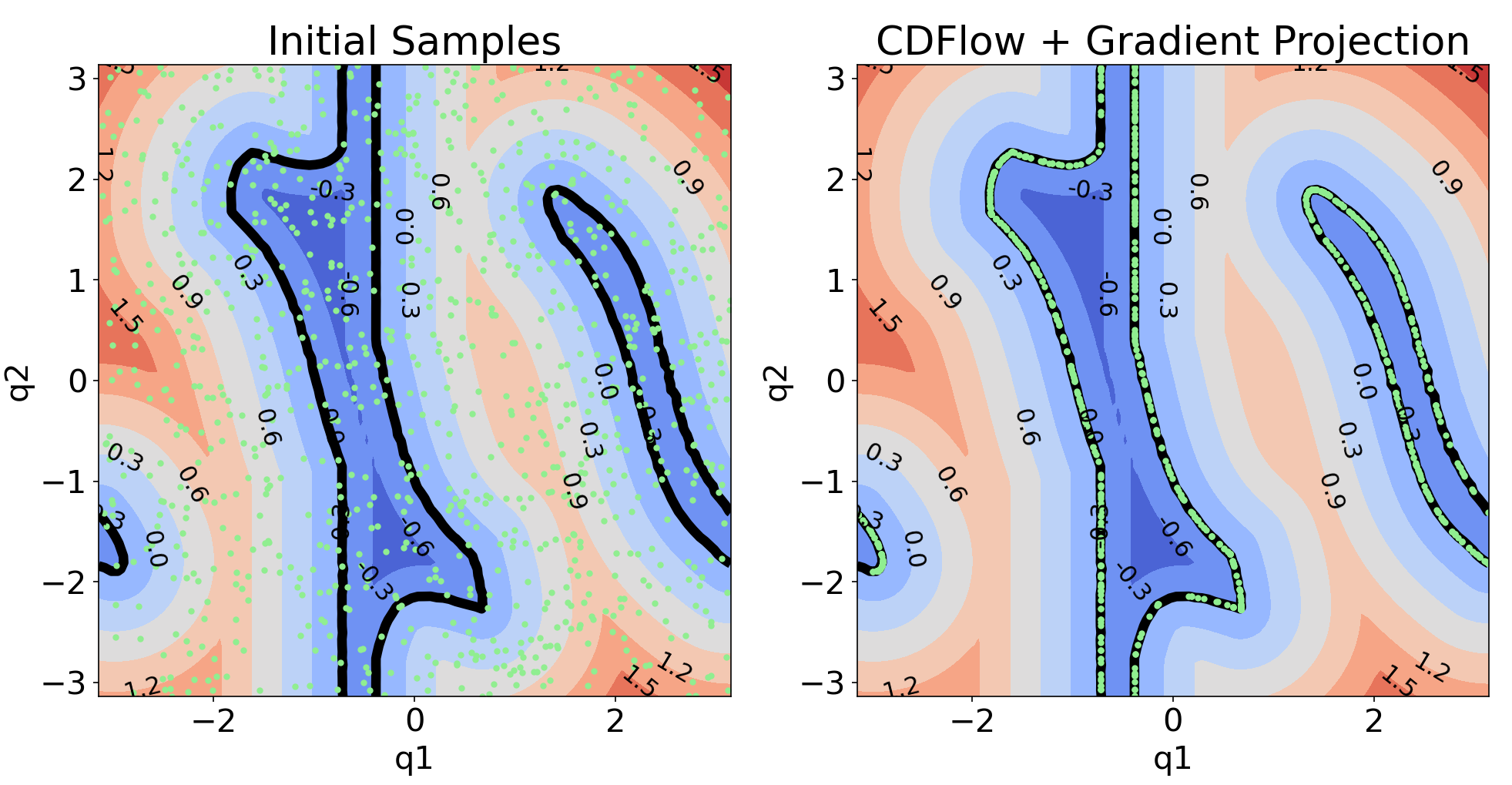}\\[2mm]
  \centering (b)
\end{minipage}\hfill
\begin{minipage}{0.2\linewidth}   % (c)
  \includegraphics[width=\linewidth]{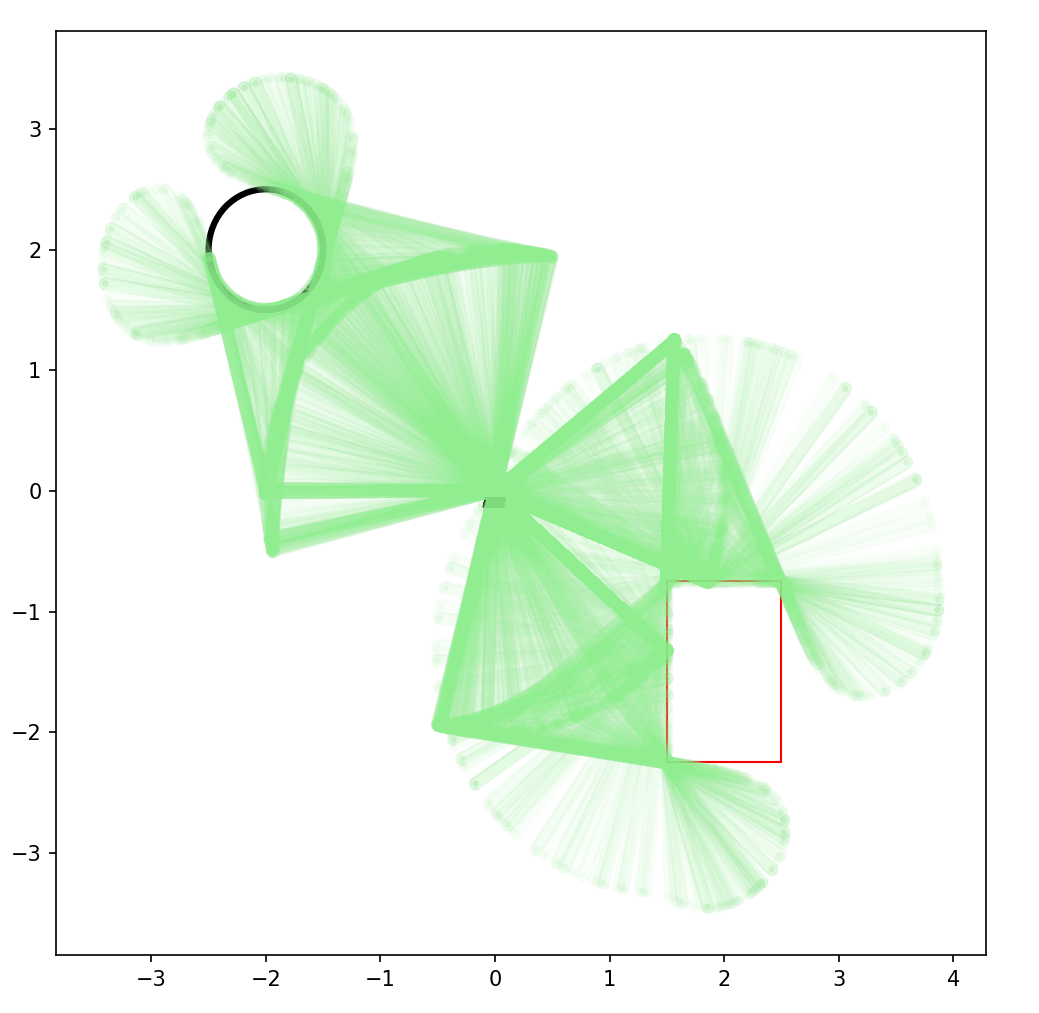}\\[2mm]
  \centering (c)
\end{minipage}
\caption{\textbf{Visualization of the core challenge motivating CDFlow: the multi-modality of minimal-distance collisions and the resulting gradient ambiguity.} 
\textbf{(a)} A PyBullet simulation showing a Franka manipulator interacting with a small sphere (radius 0.01m). For a single query configuration $\mathbf{q}$, there exists a vast, continuous manifold of collision configurations (visualized as a swept volume of the arm) that are all nearly equidistant (within a 0.0001m tolerance). This visually demonstrates the multi-modality of minimal-distance solutions.
\textbf{(b)} The consequence of this multi-modality in configuration space. When random points are projected for one step using a standard gradient-based method (as in the original CDF), the resulting directions are inconsistent and ambiguous, failing to converge to a single coherent region. This highlights the unreliability of a gradient field derived from a single nearest point.
\textbf{(c)} The task-space visualization of the zero-level-set, showing the geometric shape of the collision manifold as traced by the robot's end-effector.}
\label{fig:intro_fig}
\end{figure*}

Despite its promise, the current CDF framework exhibits two critical limitations when applied to high-DoF robots. First, by definition, CDF returns only the distance to the single nearest collision configuration. This implicitly assumes that the configuration achieving this minimum distance is unique. However, for high-DoF robots, this is often not the case; a query configuration can be equidistant to a whole manifold of different collision configurations. This multi-modality of minimal-distance solutions means that a gradient computed from any single one of these solutions is arbitrary and often misleading, leading to what we term gradient ambiguity. Second, the construction of CDF relies on optimizers searching over a sparsely sampled representation of the collision boundary. In high-dimensional spaces, this process is prone to local minima and often fails to discover the true minimal distance, resulting in an inaccurate and overly smooth distance field. This effect, which we call geometric distortion, masks sharp geometric features critical for planning. Figure~\ref{fig:intro_fig} visually illustrates these fundamental challenges.

To overcome these limitations, we propose CDFlow, a framework that models configuration-space geometry not as a static scalar field but as a continuous gradient flow. We shift the paradigm from finding a single point to learning the entire distribution of minimal-distance collision configurations, denoted as $P(\mathbf{q}'|\mathbf{q}, \text{scene})$. This explicitly acknowledges and models the multi-modal nature of the problem. Our gradient flow is then derived as the expected direction towards this distribution, providing a robust, ambiguity-free vector field for planners. To combat geometric distortion, we introduce an adaptive refinement sampling strategy that generates high-fidelity data points lying on the true minimal-distance boundary. We employ Neural Ordinary Differential Equations (Neural ODEs)~\cite{chen2018neural} to learn this complex, conditional distribution, leveraging their power as continuous normalizing flows~\cite{grathwohl2018ffjord} to generate diverse and accurate samples.  The resulting framework provides consistent and informative gradients throughout the configuration space, mitigating gradient ambiguity while faithfully representing the underlying collision geometry.

The main contributions of this paper are summarized as follows:
\begin{itemize}
    \item We propose CDFlow, a novel framework that applies Neural ODEs to learn a continuous gradient flow over the distribution of minimal-distance configurations.
    \item We present a distributional redefinition of the CDF problem, shifting from a point-to-point metric to modeling a conditional point-to-distribution relationship.
    \item We develop an adaptive refinement sampling strategy that efficiently generates higher-fidelity samples of the minimal-distance collision boundary.
    \item We validate our approach through extensive experiments, demonstrating improved efficiency, quality, and robustness compared to baseline CDF-based methods.
\end{itemize}

\begin{figure*}[t]
    \centering
    \includegraphics[width=0.9\textwidth]{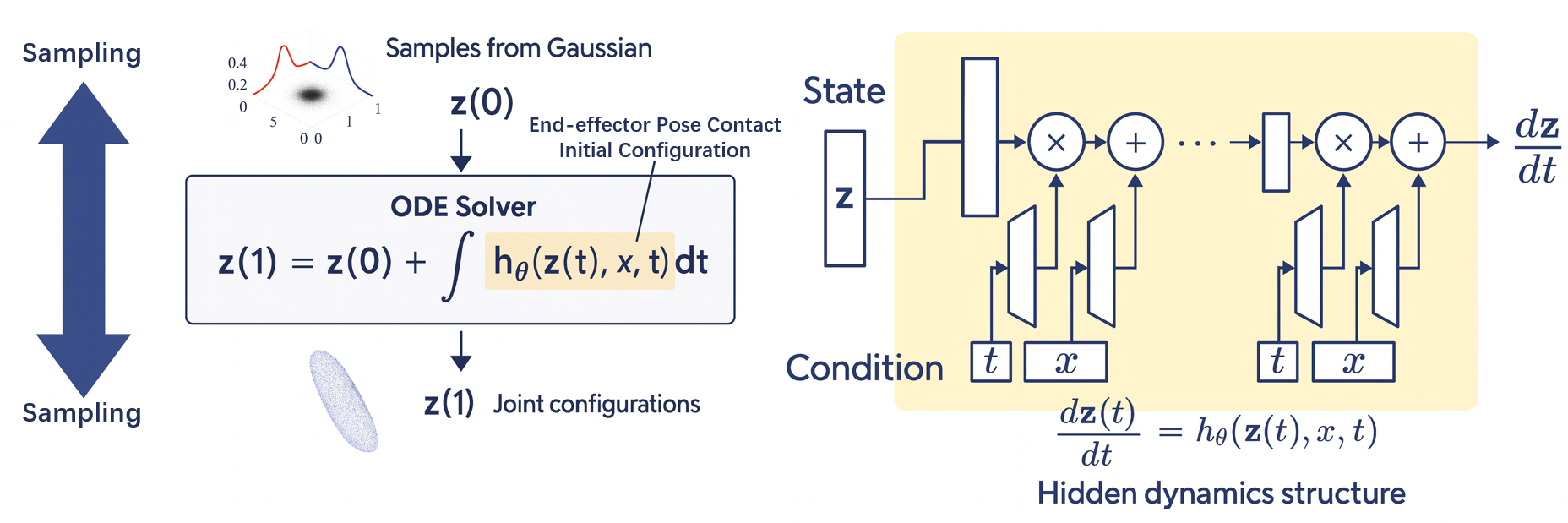} % 图片路径
    \caption{\textbf{Architecture of our conditional Continuous Normalizing Flow (CNF) for learning the collision distribution.} 
    \textbf{(Left)} The generative process of CDFlow. A sample from a simple latent distribution, $\mathbf{z}(0)$, is transformed into a sample from the multi-modal collision distribution, $\mathbf{z}(1)$, by integrating a learned, conditional vector field $h_\theta$. The invertibility of this flow allows for both efficient sampling and exact likelihood computation.
    \textbf{(Right)} The structure of the hidden dynamics network $h_\theta$. The transformation is continuously conditioned on the state $\mathbf{z}(t)$, the integration time $t$, and the input query $\mathbf{x}$ (which contains both the query configuration $\mathbf{q}$ and scene information). This architecture enables CDFlow to learn a unique and complex collision distribution for every distinct query.}
    \label{fig:model-struct}
\end{figure*}

% 高自由度 (DoF) 机器需要在复杂环境中进行安全高效的运动规划。这项任务带来了一个核心的几何挑战：生成满足所有约束的、平滑且无碰撞的轨迹。一个精确的几何表示是其关键。符号距离场 (SDFs) 已成为一个强大的工具，为规划和优化提供了连续的距离和解析梯度。
% SDF在任务空间中运行，但机器人控制发生在配置空间 (C-space) 中。配置空间距离场 (CDF) 的提出正是为了连接这一鸿沟。它直接在C空间中编码一个场景（机器人与环境）的碰撞几何。这为规划器提供了结构化的梯度，避免了昂贵的坐标变换。
% 尽管前景广阔，现有CDF框架对高自由度机器人存在着关键局限。首先，它假设最近的碰撞位形是唯一的。这很少成立。一个查询位形常常与一整个流形上的不同碰撞位形等距。这种最小距离解的多模态性创造了我们所称的梯度模糊性：从任意单个解得到的梯度都是武断且具有误导性的。其次，CDF的构建依赖于优化器在稀疏采样的碰撞边界上搜索。这个过程常常找到局部最优而非真正的最小距离。其结果是一个不准确、过度平滑的场，遭受着几何失真，掩盖了对规划至关重要的锐利特征。
% 为克服这些问题，我们提出了 CDFlow。我们的框架将C空间几何建模为一个连续的梯度流，而非一个静态的标量场。我们将范式从“寻找一个点”转变为“学习整个最小距离碰撞位形的分布”， P(q′∣q,scene)
% 。这显式地对问题的多模态特性进行建模。梯度流则是朝向该分布的期望方向，提供了一个鲁棒、无模糊的向量场。为对抗几何失真，我们引入了一种自适应精化采样策略，以发现并学习位于真实最小距离边界上的高保真度点。我们采用神经微分方程 (Neural ODEs) 来学习这个复杂的条件分布，并利用其作为连续归一化流的强大能力来生成多样且精确的样本。
% 本文的主要贡献如下：
%  CDFlow： 一个应用Neural ODEs学习最小距离位形分布上连续梯度流的新颖框架。
%  分布式的CDF公式： 从点对点的度量转变为对一个条件的点对分布关系进行建模。
%  自适应精化采样： 一种高效生成最小距离碰撞边界高保真度样本的策略。
%  实验验证： 证明了在规划成功率、轨迹质量和效率上，相比基准方法均有显著提升。

\section{Related Work}
\label{sec:related_work}

Our work, CDFlow, is situated at the confluence of continuous geometric representations, configuration space analysis, and generative modeling for robot kinematics.

\paragraph{Signed Distance Fields in Robotics.}
Signed Distance Fields (SDFs) are a cornerstone of modern geometric representation, with deep roots in computer graphics for tasks like shape encoding~\cite{park2019deepsdf} and differentiable rendering~\cite{vicini2022differentiable, mildenhall2021nerf}. The robotics community has increasingly adopted SDFs for their efficacy in providing continuous distance and gradient information. This has catalyzed progress in various domains, including gradient-based motion planning~\cite{ratliff2009chomp, zucker2013chomp}, real-time 3D mapping~\cite{izadi2011kinectfusion}, and contact-rich manipulation~\cite{driess2022learning, weng2023ngdf}. To better suit articulated robots, recent works have explored learning SDFs conditioned on joint angles~\cite{koptev2022neural} or representing the robot's swept volume~\cite{michaux2023reachability}, culminating in powerful whole-body representations for complex manipulation tasks~\cite{li2024representing}. Our work builds upon this rich body of literature but shifts the focus from task-space representations to the geometry of the configuration space itself.

\paragraph{Representations of Configuration Space.}
While powerful, the prevailing focus of SDFs in robotics has been on the task space. Consequently, actions in the configuration space (C-space) are often derived indirectly through mappings between the two spaces~\cite{Siciliano2008robotics, ratliff2018riemannian}. Traditional C-space representations frequently rely on binary collision maps~\cite{schulman2014motion, werner2023approximating}, which are foundational for classic sample-based motion planners~\cite{lavalle2001randomized, williams2017model, bhardwaj2022storm} but lack gradient information. In contrast, representing C-space geometry as a continuous distance field---a Configuration Space Distance Field (CDF)~\cite{li2024configuration}---enables the direct application of gradient-based frameworks, such as Riemannian motion policies~\cite{ratliff2018riemannian}, geometric fabrics~\cite{van2022geometric}, and other metric-aware planners~\cite{ratliff2015understanding, klein2023design}. \textbf{CDFlow advances this paradigm by addressing the critical limitations of gradient ambiguity and geometric distortion in the original CDF formulation through a novel distributional approach.}

\paragraph{Generative Models for Robot Kinematics.}
The core of our method is modeling a multi-modal distribution of configurations, aligning with recent work on generative inverse kinematics (IK). To capture the diverse solution space of redundant robots, methods have employed various generative models, from VAEs and GANs~\cite{acar2021approximating} to, more recently, normalizing flows. IKFlow~\cite{ames2022ikflow} demonstrated the power of discrete flows for generating diverse IK solutions. Our work leverages Continuous Normalizing Flows (CNFs)~\cite{chen2018neural}, following the pioneering work of NODE IK~\cite{park2022node} which first applied Neural ODEs to this problem. While related to methods that learn CNFs on explicit Riemannian manifolds~\cite{ mathieu2020riemannian}, our approach learns a distribution on a data-driven implicit manifold. \textbf{Conceptually, CDFlow extends the idea of learning a distribution of \emph{task-achieving} solutions (generative IK) to learning a distribution of \emph{minimal-distance collision} solutions, providing a richer, more robust geometric understanding for planning.}

\section{Methodology}
\label{sec:methodology}

In this section, we present our proposed framework, \textbf{CDFlow}. We begin with the necessary preliminaries, including signed distance fields (SDFs), configuration space distance fields (CDFs), and Neural ODEs as generative models. We then introduce a distributional redefinition of CDF, followed by an adaptive sampling strategy for approximating the collision distribution. Finally, we describe how a unified network, featuring a Continuous Normalizing Flow (CNF), is employed to learn the gradient flow in configuration space.

%-------------------------------------------------------------------------
\subsection{Preliminaries}

\subsubsection{Signed Distance Fields (SDFs)}
A signed distance field (SDF) is a continuous function $f_s: \mathbb{R}^3 \to \mathbb{R}$ that maps a workspace point $\mathbf{x}$ to its signed Euclidean distance from an object boundary $\partial\Omega$:
\begin{equation}
    f_s(\mathbf{x}) = \text{sign}(\mathbf{x}) \min_{\mathbf{x}' \in \partial\Omega} \|\mathbf{x} - \mathbf{x}'\|_2.
\end{equation}
Here $\text{sign}(\mathbf{x})$ is positive outside the object, negative inside, and zero on the boundary. SDFs possess two key properties critical for robotics: (i) they are differentiable almost everywhere, and (ii) their gradients have unit norm, i.e., $\|\nabla f_s(\mathbf{x})\|_2 = 1$. The gradient $\nabla f_s(\mathbf{x})$ points toward the nearest surface location, offering an optimal direction for obstacle avoidance or contact tasks.

\subsubsection{Configuration Space Distance Fields (CDFs)}
While SDFs are defined in task space, planning ultimately occurs in the high-dimensional configuration space (C-space, $\mathcal{Q}$). The configuration space distance field (CDF)~ encodes collision geometry directly in $\mathcal{Q}$. Formally, for a configuration $\mathbf{q} \in \mathcal{Q}$, the CDF is
\begin{equation}
    f_c(\mathbf{q}) = \min_{\mathbf{q}' \in \mathcal{Q}_c} \|\mathbf{q} - \mathbf{q}'\|_2,
\end{equation}
where $\mathcal{Q}_c$ denotes the set of colliding configurations. This $\min$ operator effectively imposes a \emph{uni-modal assumption}: it selects a single nearest collision point even when multiple equidistant collisions exist. Moreover, since enumerating $\mathcal{Q}_c$ is infeasible, CDFs must be built from sparse optimization-derived samples, which can oversmooth the geometry and distort sharp boundaries.

\subsubsection{Neural ODEs as Generative Models}
Normalizing flows (NFs) are generative models that learn invertible mappings from a simple prior distribution $P_z$ (e.g., Gaussian) to a complex data distribution $P_x$. Continuous normalizing flows (CNFs extend this to continuous transformations via Neural ODEs. A CNF is defined by solving an ODE with neural dynamics:
\begin{equation}
    \frac{d\mathbf{z}(t)}{dt} = g(\mathbf{z}(t), t; \theta),
\end{equation}
where $g$ is a neural network parameterized by $\theta$. The change of log-density follows the instantaneous change of variables:
\begin{equation}
    \frac{d \log p(\mathbf{z}(t))}{dt} = -\text{Tr}\!\left(\frac{\partial g}{\partial \mathbf{z}(t)}\right).
\end{equation}
This formulation enables modeling high-dimensional, multi-modal distributions with parameter efficiency, making CNFs suitable for approximating complex collision distributions in configuration space.

%-------------------------------------------------------------------------
\subsection{Distributional Redefinition of CDF}
\label{sec:dist_cdf}

To overcome the uni-modal limitation of the original CDF, we redefine its core outputs from a distributional perspective. Given a query configuration $\mathbf{q}$, we consider the entire set of minimal-distance collision configurations, denoted as $\mathcal{Q}^*(\mathbf{q}, \text{scene})$, not just a single point from it. We model this set as the support of a conditional probability distribution $P(\mathbf{q}'|\mathbf{q}, \text{scene})$.

From this distribution, we define the \textbf{CDFlow distance} and \textbf{gradient field} as expectations over this distribution:
\begin{equation}
    f_{\text{flow}}(\mathbf{q}) = \mathbb{E}_{\mathbf{q}' \sim P(\mathbf{q}'|\mathbf{q}, \text{scene})} [\|\mathbf{q} - \mathbf{q}'\|_2].
    \label{eq:cdflow_dist_revised}
\end{equation}
\begin{equation}
    \nabla f_{\text{flow}}(\mathbf{q}) = \mathbb{E}_{\mathbf{q}' \sim P(\mathbf{q}'|\mathbf{q}, \text{scene})} \left[ \frac{\mathbf{q} - \mathbf{q}'}{\|\mathbf{q} - \mathbf{q}'\|_2} \right].
    \label{eq:cdflow_grad_revised}
\end{equation}
This expectation-based formulation fundamentally resolves the ambiguity of the $\min$ operator inherent in the original CDF. It provides a smooth and consistent descent direction by integrating information from all minimal-distance collision modes, yielding a far more robust signal for downstream planners.

%-------------------------------------------------------------------------
\subsection{Approximating the Distribution via Adaptive Sampling}
\label{sec:sampling}

Since the distribution $P(\mathbf{q}'|\mathbf{q}, \text{scene})$ is analytically intractable, we approximate its support set $S(\mathbf{q}, \text{scene})$ for a given query $\mathbf{q}$ through an adaptive refinement sampling strategy. The procedure consists of two stages:  

\textbf{(i) Global Exploration.} Starting from random seeds, we repeatedly run an optimizer (e.g., L-BFGS) to find a diverse set of locally optimal solutions for the minimal-distance problem, aiming to identify distinct modes of $\mathcal{Q}^*(\mathbf{q}, \text{scene})$.  

\textbf{(ii) Local Refinement.} For each high-quality solution $\mathbf{q}'_i$ discovered, we densely resample within a local neighborhood $\mathcal{N}(\mathbf{q}'_i, \epsilon)$ and re-run a local optimization. This efficiently discovers other points on the same minimal-distance manifold, improving local geometric accuracy.  

\begin{algorithm}[t]
\caption{Adaptive Refinement Sampling}
\label{alg:sampling}
\begin{algorithmic}[1]
\State \textbf{Input:} Robot model, scene, query config $\mathbf{q}$, global seeds $N_g$, local seeds $N_l$, radius $\epsilon$
\State \textbf{Output:} High-fidelity collision sample set $S$
\State $S \gets \emptyset$; $S_{global} \gets \emptyset$
\For{$i = 1$ to $N_g$}
    \State $\mathbf{q}_{\text{init}} \gets \text{RandomSample}(\mathcal{Q})$
    \State $\mathbf{q}' \gets \text{OptimizeToBoundary}(\mathbf{q}_{\text{init}}, \mathbf{q})$
    \State $S_{global} \gets S_{global} \cup \{\mathbf{q}'\}$
\EndFor
\State Find minimum distance $d_{\min}$ from $S_{global}$
\State $S \gets \{ \mathbf{q}' \in S_{global} \mid \|\mathbf{q}-\mathbf{q}'\|_2 \approx d_{\min} \}$
\For{each $\mathbf{q}'_i \in S$}
    \For{$j = 1$ to $N_l$}
        \State $\mathbf{q}_{\text{init}} \gets \text{RandomSampleBall}(\mathbf{q}'_i, \epsilon)$
        \State $\mathbf{q}'' \gets \text{OptimizeToBoundary}(\mathbf{q}_{\text{init}}, \mathbf{q})$
        \If{$\|\mathbf{q}-\mathbf{q}''\|_2 \approx d_{\min}$}
            \State $S \gets S \cup \{\mathbf{q}''\}$
        \EndIf
    \EndFor
\EndFor
\State \Return $S$
\end{algorithmic}
\end{algorithm}

The resulting set $S(\mathbf{q}, \text{scene})$ forms an empirical approximation of the support of $P(\mathbf{q}'|\mathbf{q}, \text{scene})$.

%-------------------------------------------------------------------------
\subsection{Learning the Gradient Flow via a Unified Network}
\label{sec:learning}

Our objective is to train a single, unified network $F_\theta$ that performs two coupled tasks: implicitly learning the conditional distribution $P(\mathbf{q}'|\mathbf{q}, \text{scene})$ and explicitly predicting a geometrically consistent distance field. The network is trained end-to-end with a hybrid objective. For a given input $(\mathbf{q}, \text{scene})$, the network $F_\theta$ provides two outputs:
\begin{enumerate}
    \item A generative mapping from a prior noise vector $\mathbf{z}_0$ to a collision sample $\hat{\mathbf{q}}'$, realized via a Continuous Normalizing Flow (CNF) block.
    \item A direct regression of the scalar distance value, $\hat{d}$, from which the gradient $\nabla_{\mathbf{q}} \hat{d}$ is obtained via automatic differentiation.
\end{enumerate}

The architecture of our network, implemented as a conditional Continuous Normalizing Flow, is detailed in Figure~\ref{fig:model-struct}.

Crucially, both the CNF dynamics and the distance regression head are conditioned on the input and share parameters, ensuring that the learned distribution informs the predicted gradient field.

\paragraph{Training Objective.}
For each training instance, consisting of a query $\mathbf{q}$ and its corresponding adaptively sampled collision set $S(\mathbf{q}, \text{scene})$, we define the following losses:

\textbf{(i) Distributional Alignment via CNF.}
The generative component of our network is trained to maximize the likelihood of the sampled collision configurations:
\begin{equation}
    \mathcal{L}_{\text{NLL}} = - \frac{1}{|S|} \sum_{\mathbf{q}' \in S} \log p_\theta(\mathbf{q}'|\mathbf{q}, \text{scene}).
\end{equation}
This loss compels the network to capture the underlying multi-modal structure of the minimal-distance solutions.

\textbf{(ii) Geometric Consistency with Purified Targets.}
The regressive component of the network is supervised using robust, expectation-based ground truth signals computed from the high-fidelity sample set $S$:
\begin{align}
    y_d(\mathbf{q}) &= \frac{1}{|S|} \sum_{\mathbf{q}' \in S} \|\mathbf{q} - \mathbf{q}'\|_2, \label{eq:gt_dist} \\
    y_g(\mathbf{q}) &= \frac{1}{|S|} \sum_{\mathbf{q}' \in S} \frac{\mathbf{q} - \mathbf{q}'}{\|\mathbf{q} - \mathbf{q}'\|_2}. \label{eq:gt_grad}
\end{align}
We then supervise the network's direct distance prediction $\hat{d}$ and its gradient $\nabla_{\mathbf{q}} \hat{d}$ against these purified targets:
\begin{align}
    \mathcal{L}_{\text{dist}} &= \|\hat{d} - y_d(\mathbf{q})\|_2^2, \\
    \mathcal{L}_{\text{grad}} &= 1 - \frac{(\nabla_{\mathbf{q}} \hat{d}) \cdot y_g(\mathbf{q})}{\|\nabla_{\mathbf{q}} \hat{d}\|_2 \|y_g(\mathbf{q})\|_2}.
\end{align}
We also include the Eikonal and Tension losses as structure-promoting regularizers on the predicted field:
\begin{align}
    \mathcal{L}_{\text{eik}}  &= \Big|\|\nabla_{\mathbf{q}} \hat{d}\|_2 - 1\Big|, \\
    \mathcal{L}_{\text{ten}}  &= \|\nabla^2_{\mathbf{q}} \hat{d}\|_F^2.
\end{align}

\paragraph{Overall Loss.}
The unified network $F_\theta$ is trained by minimizing a weighted sum of all components:
\begin{equation}
    \mathcal{L}_{\text{total}} = 
    \lambda_1 \mathcal{L}_{\text{NLL}} + 
    \lambda_2 \mathcal{L}_{\text{dist}} + 
    \lambda_3 \mathcal{L}_{\text{grad}} + 
    \lambda_4 \mathcal{L}_{\text{eik}} + 
    \lambda_5 \mathcal{L}_{\text{ten}}.
\end{equation}
This hybrid objective ensures that the expressive power of the generative model in capturing the collision distribution directly regularizes and improves the quality of the predicted gradient field.

\paragraph{Inference.}
Once trained, we can leverage either output of the network. For the most accurate, ambiguity-free gradient, we use the generative component to perform Monte Carlo approximation as described in Sec.~\ref{sec:dist_cdf}. For faster queries, we can directly use the gradient $\nabla_{\mathbf{q}} \hat{d}$ from the network's regressive output, which serves as an efficient, learned approximation of the expected gradient.

\section{Experiments}
\label{sec:experiments}
\begin{figure*}[t]
    \centering
    % Placeholder for a 1x3 or 2x2 figure
    \includegraphics[width=\linewidth]{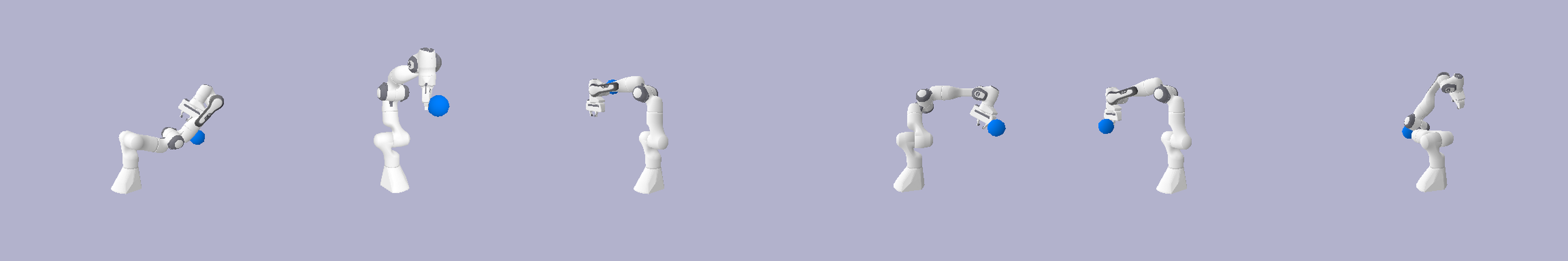}
    \caption{Qualitative comparison on a challenging button-pressing task. (a) The setup, showing the robot and a difficult-to-reach button. (b) The CDF-based method fails due to an ambiguous gradient, resulting in an oscillating trajectory. (c) Our CDFlow provides a smooth, unambiguous gradient, enabling a successful one-shot projection.}
    \label{fig:wb_ik_sim}
\end{figure*}

\begin{figure*}[!t]
    \begin{center}
        \begin{tabular}{cccc}
            \centering
            \includegraphics[width =0.23\linewidth
            ]{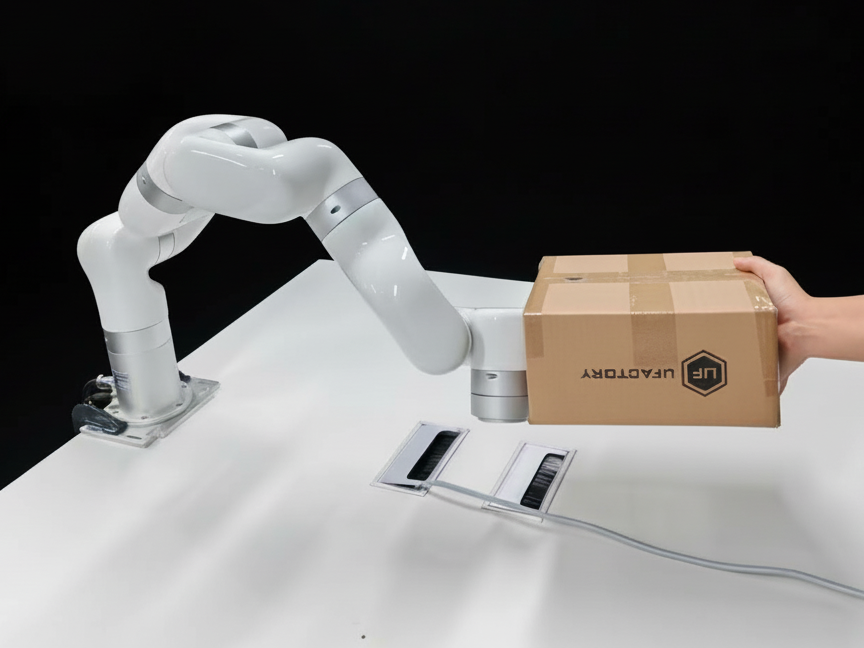}   &
            \includegraphics[width =0.23\linewidth]{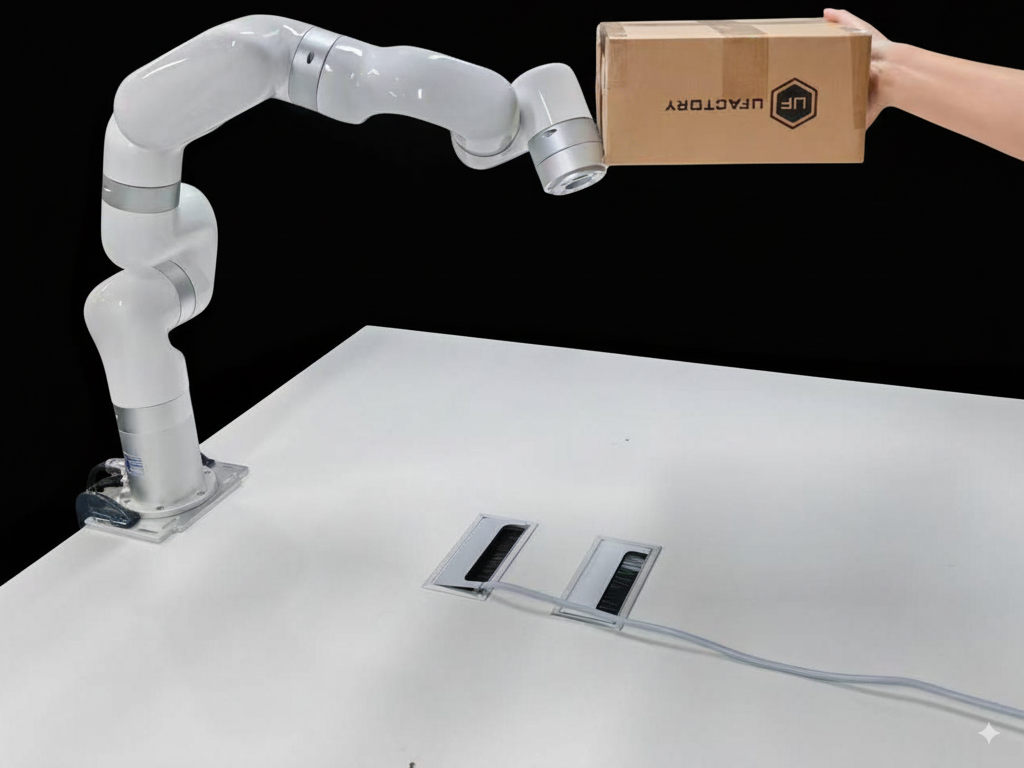}  &

            \includegraphics[width =0.23\linewidth]{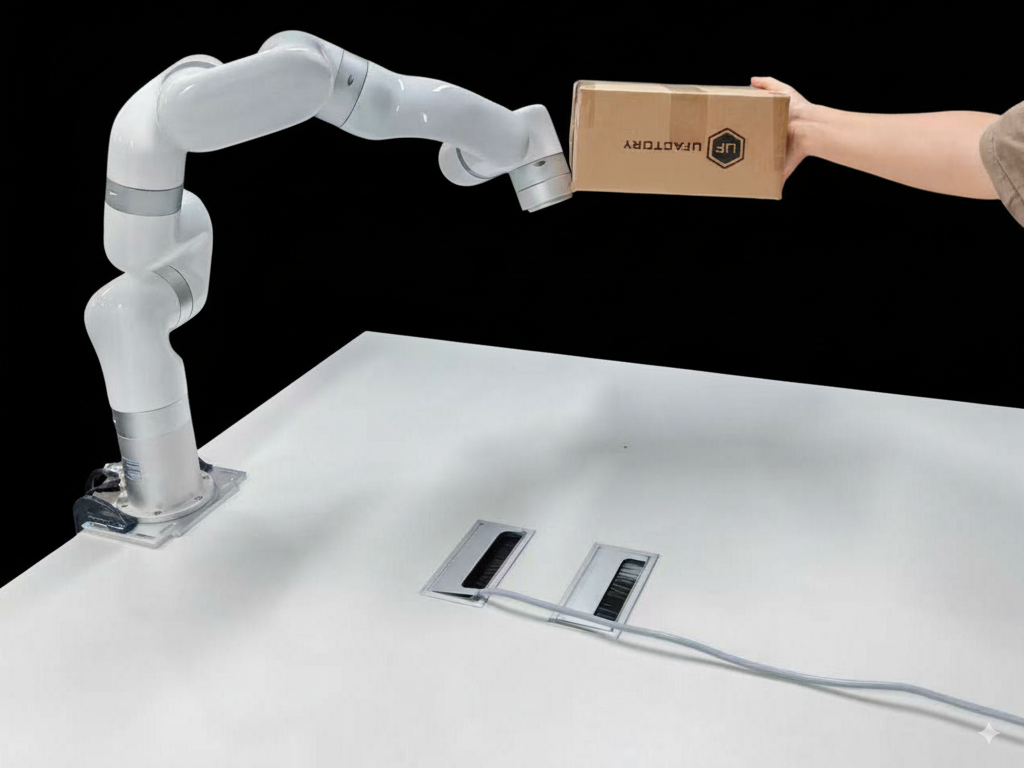}  &
            \includegraphics[width =0.23\linewidth]{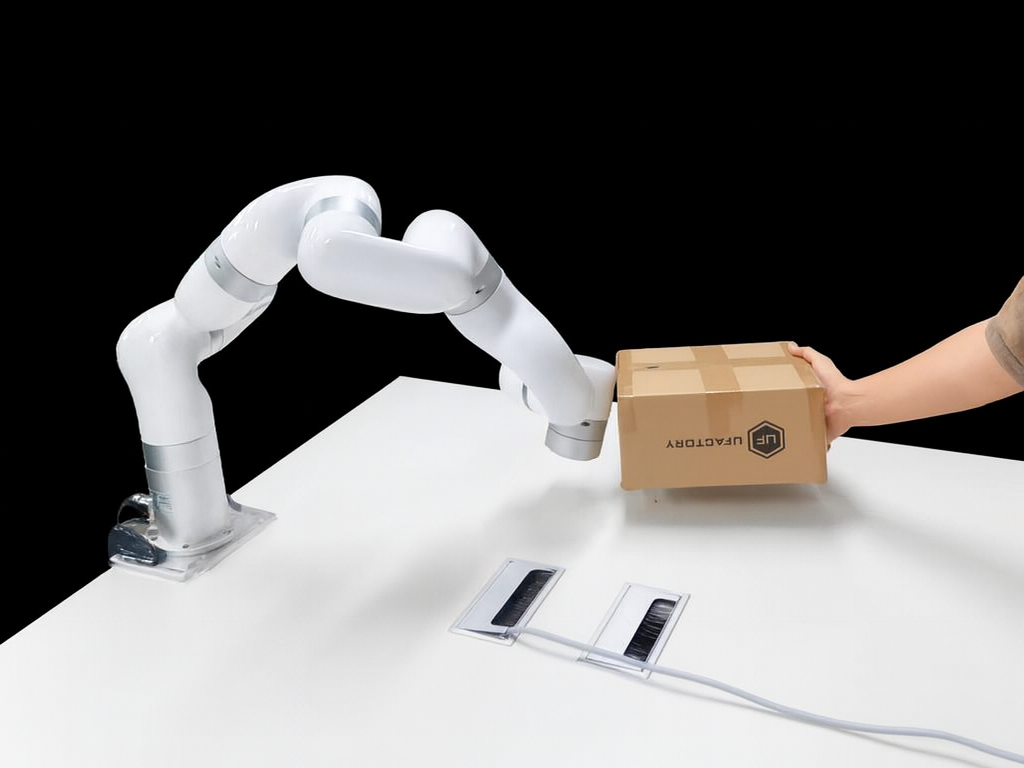}  \\
        \end{tabular}
    \caption{Snapshots from a real-world human-robot collaborative carrying experiment with a 7-DoF xArm robot, enabled by CDFlow. 
    (Left to Right) The robot perceives the box's pose in real-time using a depth camera. It then performs whole-body projection using CDFlow's robust gradient to establish contact. Finally, it coordinates its motion with a human partner to collaboratively carry the box across the workspace. The smooth, stable, and reactive motions generated by our planner highlight CDFlow's potential for enabling complex and safe physical human-robot interaction in unstructured environments.}
        \label{fig:wb_ik}
    \end{center}
\end{figure*}

\begin{table*}[!t]
    \begin{center}
        \begin{tabular}{ccccc}
            \centering
            \includegraphics[width =0.18\linewidth]{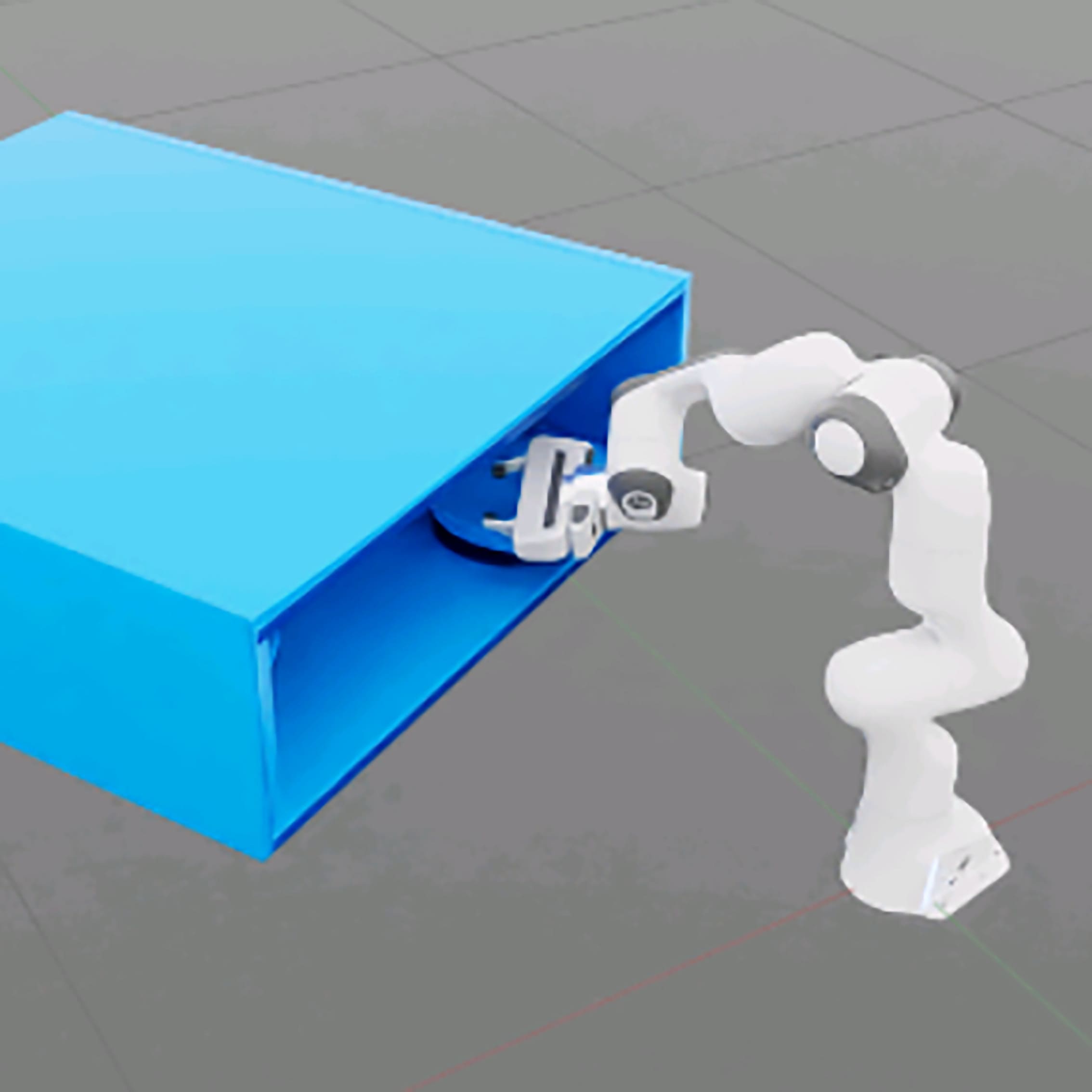}   &
            \includegraphics[width =0.18\linewidth]{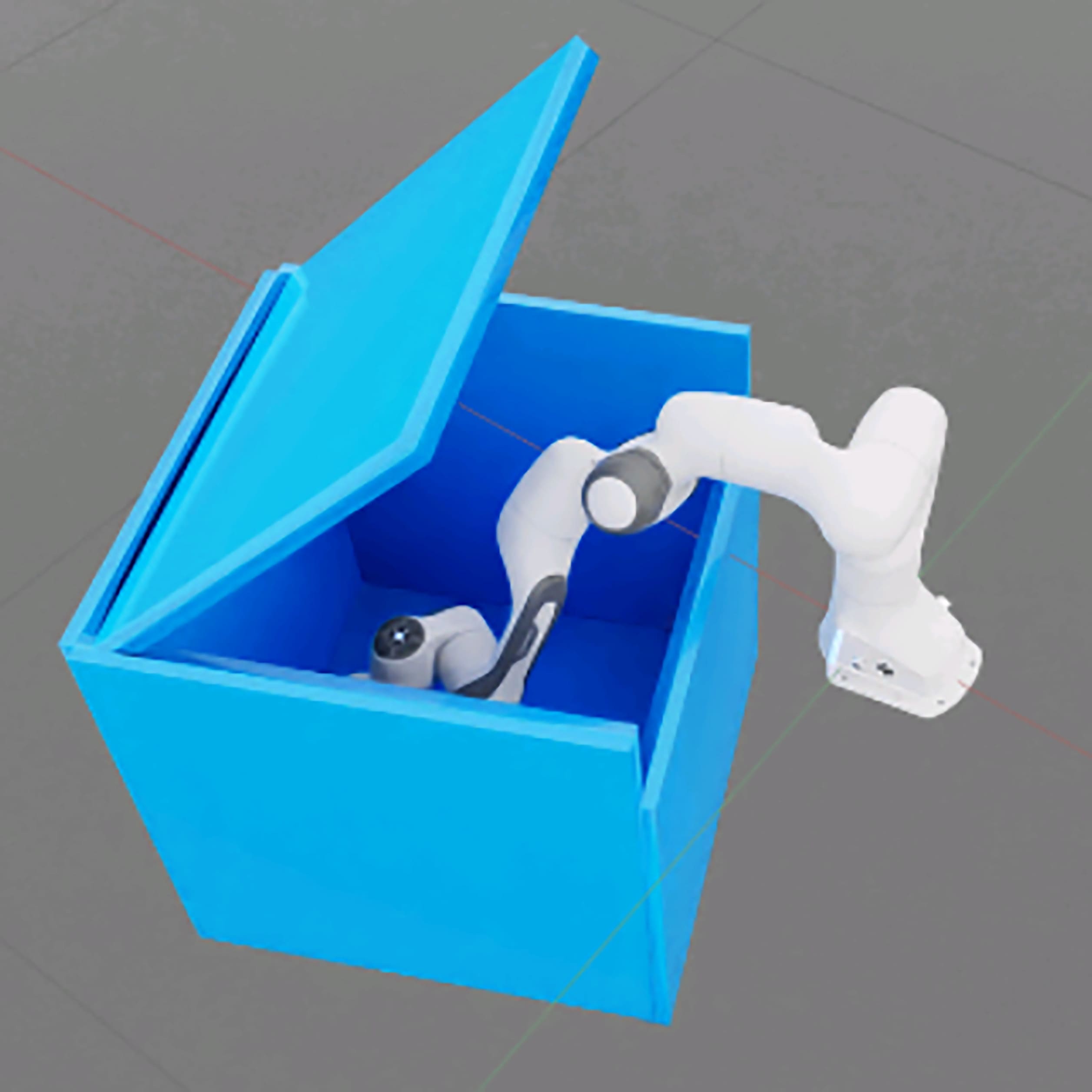}  &

            \includegraphics[width =0.18\linewidth]{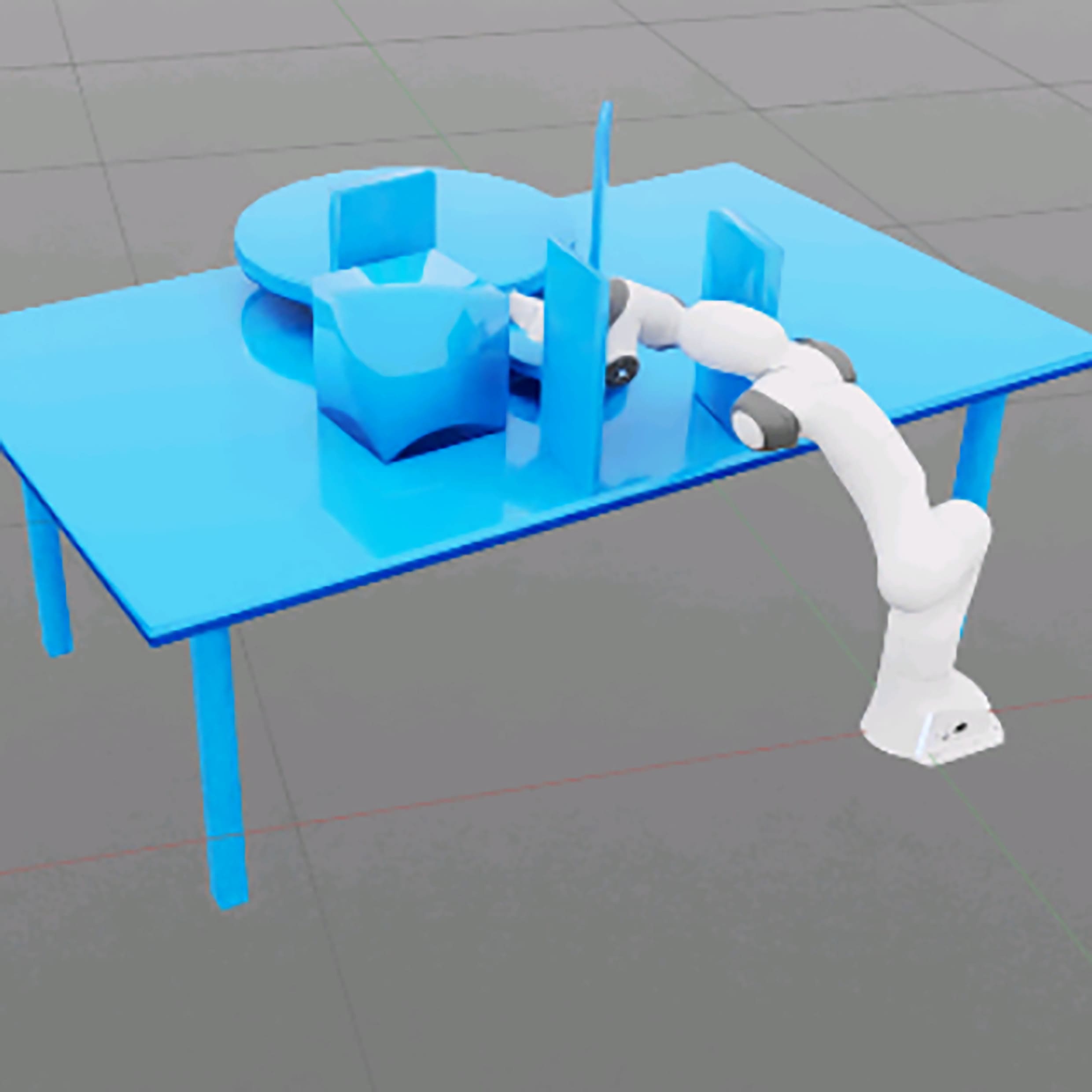}  &
            \includegraphics[width =0.18\linewidth]{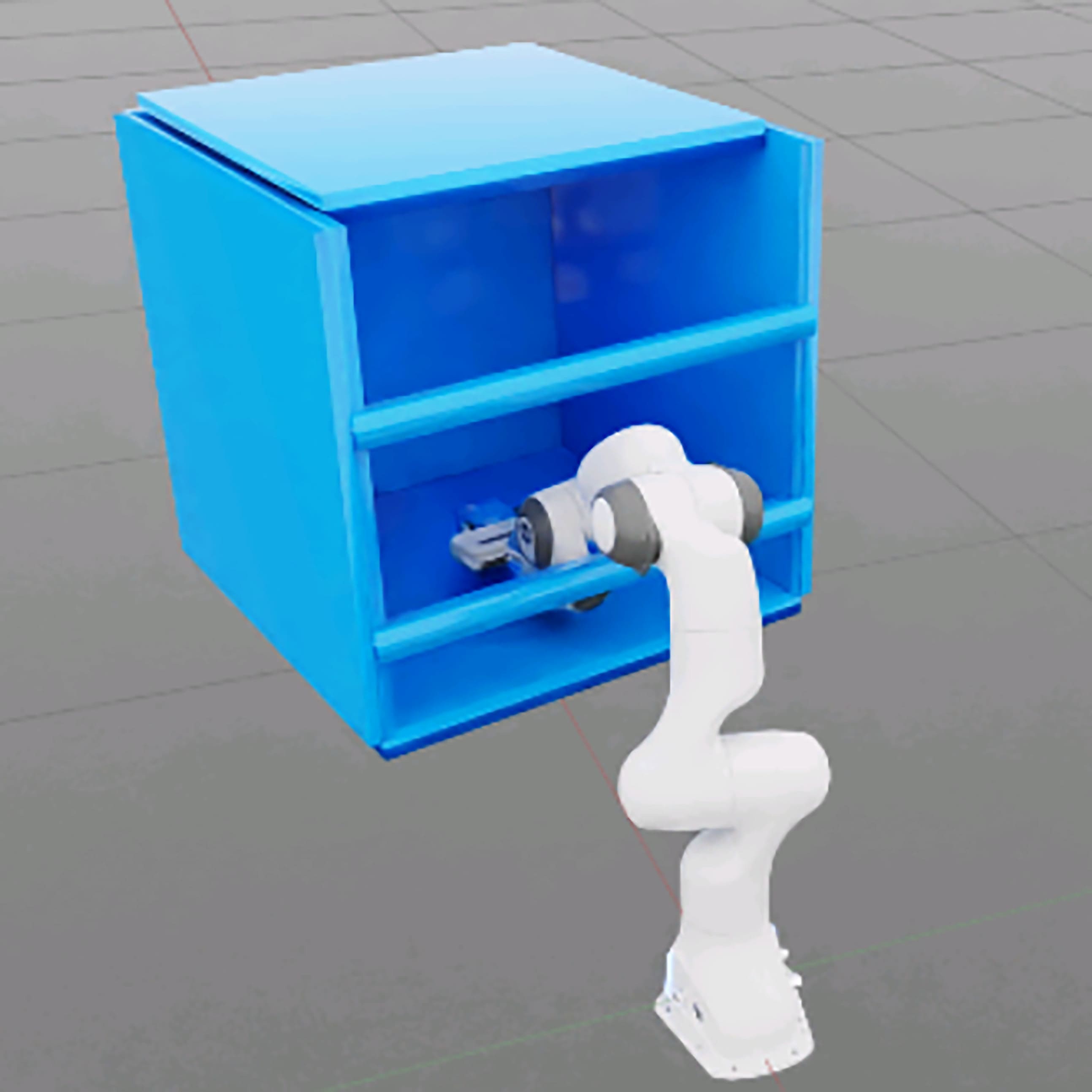}  &

            \includegraphics[width =0.18\linewidth]{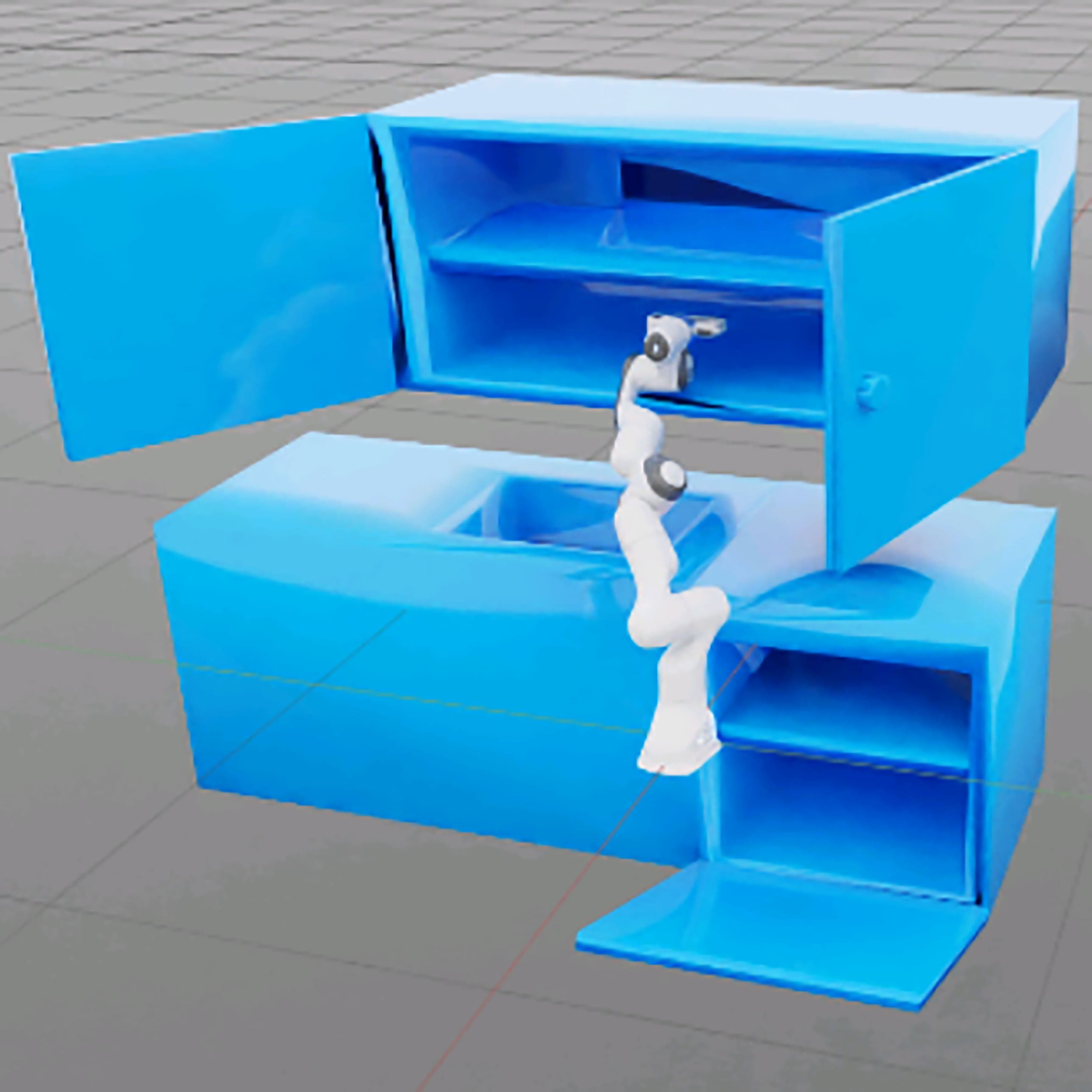}  \\
        \end{tabular}
        \captionof{figure}{A sequence of snapshots from the human-robot collaborative box-carrying task. Our CDFlow-based planner generates smooth, coordinated whole-body motions to follow the human partner.
        }
        % \vspace{-10mm}
        \label{fig:motion_bench}
    \end{center}
\end{table*}

In this section, we conduct a series of rigorous experiments to validate the effectiveness of our proposed CDFlow framework. Our evaluation is designed to answer three core questions: 
(1) Do the reliable gradients provided by CDFlow significantly improve the success rate of projection-based whole-body planning tasks, especially in scenarios designed to induce ambiguity?
(2) Can our method be extended to complex human-robot collaboration scenarios that demand robust, real-time planning?
(3) How does CDFlow perform against state-of-the-art methods on a standardized motion planning benchmark?
All experiments were conducted on a workstation equipped with an NVIDIA RTX 4090 GPU and an Intel Core i9-13900K CPU.

%-------------------------------------------------------------------------
\subsection{Implementation Details}
\label{sec:implementation}

The CDFlow model in all our experiments is based on a unified network architecture, comprising a generative component (CNF) and a regressive component (MLP), which share parameters and are trained end-to-end.

The core of the generative component is a Neural ODE-based Continuous Normalizing Flow (CNF). To ensure its expressive power, the hidden dynamics network consists of four fully connected layers with a width of 1024 and ReLU activation functions, a design inspired by successful applications in modeling complex kinematic distributions.

For the regressive component, which directly predicts the distance field, we adopt an architecture identical to our baseline to ensure a fair comparison. It consists of a 5-layer MLP. Following their approach, we also employ positional encoding~\cite{mildenhall2021nerf} on the input features to better capture high-frequency geometric details.

Our training protocol also mirrors the CDF baseline. We randomly sample $b_1=4000$ query points and $b_2=100$ configurations for each training sample. The network is trained for $50,000$ epochs using the Adam optimizer with an initial learning rate of $0.001$, which is decayed by a factor of $0.5$. The entire training process takes approximately $2$ hours on a single NVIDIA RTX 4090 GPU. The same trained model is used across all subsequent experiments.

%-------------------------------------------------------------------------
\subsection{Whole-Body Button Pressing}
\label{sec:exp_button_press}

\paragraph{Objective}
This experiment is designed to directly confront and quantify the core motivation of our work: to validate that the expected gradient from CDFlow provides a decisive advantage over the single-point gradient of traditional CDFs in the presence of gradient ambiguity caused by multi-modal solutions and kinematic singularities. This serves as a direct stress test of the gradient field quality.

\paragraph{Setup}
The experiment is conducted in the PyBullet physics simulator. A 7-DoF Franka Emika Panda robot is tasked with touching a small sphere (radius 0.01m, the "button"), which is sequentially placed at 1000 random 3D locations in front of it. 

To precisely evaluate robustness under extreme conditions, we deliberately engineer a subset of these locations to be challenging. These fall into two categories:
\begin{enumerate}
    \item \textbf{Close-Proximity and Occlusion Scenarios:} Some spheres are placed very close to the robot's base or torso. In these cases, the robot cannot easily use its end-effector and must perform complex whole-body coordination, using its elbow or upper arm links to make contact. Such scenarios naturally create a large manifold of diverse minimal-distance solutions, making them ideal for testing multi-modality handling.
    \item \textbf{Near-Singularity Scenarios:} Other spheres are positioned at the edge of the workspace, requiring the arm to be fully extended. To reach these targets, the robot must adopt configurations near kinematic singularities. In such configurations, the gradients from Jacobian-based methods and single-point CDFs become highly unstable and ambiguous.
\end{enumerate}
The task requires the robot to find a configuration that allows \emph{any part} of its body to touch the sphere. This is achieved via a pure projection-based method: at each step, the model receives the current scene (sphere location) and robot configuration $\mathbf{q}$ and outputs a gradient. The robot is projected one step along this direction. If contact is not made, the process repeats (multi-step projection). A trial is considered a failure if contact is not established within 3 seconds, after which the sphere moves to the next location. Upon successful contact, the number of steps is recorded.

\paragraph{Baselines}
We compare against two representative baselines:
\begin{itemize}
    \item \textbf{SDF-based Projection}: A classic baseline that computes gradients based on the task-space SDF of the \textbf{end-effector only}. It represents traditional planning that neglects whole-body coordination.
    \item \textbf{CDF (Li et al.)}: Our direct baseline. We strictly follow its original formulation, using its provided gradient for projection. We hypothesize that this method will frequently encounter gradient ambiguity in our challenging scenarios.
\end{itemize}

\paragraph{Results and Analysis}
The quantitative results are summarized in Table~\ref{tab:ik_comparison}. CDFlow demonstrates a dominant performance, achieving a near-perfect success rate of 97.5\%, a stark contrast to the 78.6\% of the original CDF and 43.2\% of the SDF-based method. This result strongly validates our core thesis: in challenging scenarios rich with multi-modality and singularities, traditional methods indeed suffer from severe gradient ambiguity, and the expectation-based distributional gradient of CDFlow fundamentally resolves this issue.

A deeper failure case analysis reveals the source of these performance gaps. Most failures of the SDF-based method occurred in close-proximity scenarios where non-end-effector contact was required. The failures of the CDF method were highly concentrated in our engineered challenging locations. In these cases, its single-point gradient often caused the projection to oscillate between conflicting descent directions, failing to converge within the 3-second time limit.

Regarding efficiency, CDFlow (0.81 ms) is nearly 20 times faster in query time than CDF (15.7ms). This is due to our architectural advantage: CDFlow obtains the expected gradient via a single, parallelized forward pass and Monte Carlo sampling, whereas the original CDF requires running an expensive, iterative online optimization to find the nearest point at every query. More importantly, the lower average projection steps required by CDFlow (1.2) compared to CDF (2.3) clearly indicates a higher-quality gradient. Our gradient is more informative, guiding the robot more directly towards a solution and reducing unnecessary iterations.

The qualitative comparison in Figure~\ref{fig:wb_ik_sim} further supports this analysis. In a typical challenging case, the ambiguous gradient from CDF leads to an ineffective oscillatory motion and failure. In contrast, CDFlow provides a smooth, unambiguous gradient from the start, guiding the entire arm into a coordinated posture for a one-shot successful contact.

\begin{table}[h]
\centering
\caption{Quantitative comparison on the whole-body task over 1000 trials.}
\label{tab:ik_comparison}
\resizebox{\linewidth}{!}{
\begin{tabular}{lccc}
\toprule
\textbf{Method} & \textbf{Success Rate} $\uparrow$ & \textbf{Avg. Steps} $\downarrow$ & \textbf{Query Time (ms)} $\downarrow$ \\
\midrule
SDF-based & 43.2\% & 4.1 & 2.13 \\
CDF (Li et al.) & 78.6\% & 2.3 & 1.57 \\
\textbf{CDFlow (Ours)} & \textbf{97.5\%} & \textbf{1.2} &\textbf{0.81}  \\
\bottomrule
\end{tabular}}
\end{table}

%-------------------------------------------------------------------------
\subsection{Application: Human-Robot Collaborative Box Carrying}

\begin{table*}[!t]
\centering
\begin{threeparttable}
\caption{Motion-planning comparison using gradients of SDF, CDF, and CDFlow.}
\label{tab:MP_bench}
\small
\setlength{\tabcolsep}{4pt}

% 用 tabular* 铺满 \linewidth；无竖线；第一列算法名 + 12 个数值列
\begin{tabular*}{\linewidth}{@{\extracolsep{\fill}} l
  S[table-format=3.0] S[table-format=1.2] S[table-format=3.0]
  S[table-format=3.0] S[table-format=1.2] S[table-format=3.0]
  S[table-format=3.0] S[table-format=1.2] S[table-format=3.0]
  S[table-format=3.0] S[table-format=1.2] S[table-format=3.0]}
\toprule
\multirow{2}{*}{Method}
& \multicolumn{3}{c}{\textbf{SDF}}
& \multicolumn{3}{c}{\textbf{CDF}}
& \multicolumn{3}{c}{\textbf{CDFlow}} \\
\cmidrule(lr){2-4}\cmidrule(lr){5-7}\cmidrule(lr){8-10}\cmidrule(lr){11-13}
& {\makecell{Succ.\\(\si{\percent})}}
& {\makecell{Track.\\(\si{rad})}}
& {\makecell{Opt Steps}}
& {\makecell{Succ.\\(\si{\percent})}}
& {\makecell{Track.\\(\si{rad})}}
& {\makecell{Opt Steps}}
& {\makecell{Succ.\\(\si{\percent})}}
& {\makecell{Track.\\(\si{rad})}}
& {\makecell{Opt Steps}} \\
\midrule
IPOPT      & 22 & 0.19 & 237 & 34 & 0.15 & 134 & 87 & 0.07 & 89  \\
QRQP       & 13 & 0.18 & 212 & 32 & 0.14 & 129 & 84 & 0.08 & 92  \\
OSQP       & 19 & 0.19 & 209 & 31 & 0.14 & 171 & 85 & 0.09 & 99  \\
qpOASES    & 22 & 0.16 & 194 & 37 & 0.11 & 133 & 88 & 0.05 & 85  \\

\bottomrule
\end{tabular*}

\begin{tablenotes}[flushleft]
\footnotesize
\item Succ.= success rate; Track.= average $L_{2}$ norm of the joint-space error, i.e., $\|\mathbf{q}_{\text{final}} - \mathbf{q}_{\text{target}}\|_{2}$; Opt Steps = average optimization steps.

\end{tablenotes}
\end{threeparttable}
\end{table*}

\paragraph{Objective}
To demonstrate the applicability and robustness of CDFlow beyond simulation, we designed a real-world physical human-robot interaction (pHRI) task. This experiment aims to validate whether the high-quality gradients from CDFlow can enable a real robot to perform complex, multi-point coordination tasks in real-time, reacting to dynamic changes in the environment.

\paragraph{Setup}
The experiment is conducted on a physical 7-DoF xArm7 robot mounted on a workbench. A large box is placed on the opposite side of the table. A human collaborator stands ready to lift one side of the box. The robot's task is to autonomously approach, establish a firm whole-body contact with its side of the box, and then coordinate its movements with the human to carry it.

This task poses significant real-world challenges:
\begin{enumerate}
    \item \textbf{Real-Time Perception and Planning:} The box's precise 6D pose is perceived in real-time using an RGB-D camera. The planner must continuously process this noisy, high-dimensional sensory input and generate motion commands at a high frequency.
    \item \textbf{Whole-Body Contact Planning:} Unlike simple grasping, this task requires the robot to find a stable contact configuration using its forearm or upper arm, which involves planning in its full 7-dimensional configuration space under task constraints.
    \item \textbf{Dynamic Coordination:} Once the box is lifted, the robot enters a dynamic coordination phase where it must compliantly follow the human's lead. This demands a continuous stream of stable and smooth motion plans to avoid jerky movements or losing contact.
\end{enumerate}

\paragraph{Role of CDFlow and Qualitative Results}
We use CDFlow as the core engine for a reactive, projection-based planner. The real-time pose of the box, provided by the depth camera, serves as the dynamic goal for our model. CDFlow then computes a robust, ambiguity-free gradient, which is used to project the robot's current configuration towards a valid contact state. 

As demonstrated in the experimental sequence in Figure~\ref{fig:wb_ik}, our CDFlow-based planner successfully enables the xArm7 to perform the entire task. The robot first generates a smooth, direct trajectory to make contact with the box. Subsequently, it seamlessly coordinates with the human partner, exhibiting stable and compliant carrying motions. This qualitative result is a strong testament to CDFlow's capabilities. The ability to generate high-quality, whole-body gradients from real-world sensor data in real-time is a key enabler for complex pHRI tasks. It shows that CDFlow is not just a theoretical improvement but a practical tool for building more intelligent and capable robotic systems for unstructured environments.

% \begin{figure}[t]
%     \centering
%     % Placeholder for a figure with a sequence of images
%     % \includegraphics[width=\linewidth]{figures/hri_sequence.png}
%     \caption{A sequence of snapshots from the human-robot collaborative box-carrying task. Our CDFlow-based planner generates smooth, coordinated whole-body motions to follow the human partner.}
%     \label{fig:hri}
% \end{figure}

%-------------------------------------------------------------------------
\subsection{Benchmark: MotionBenchMaker}

\paragraph{Objective}
The objective of this experiment is to comprehensively evaluate the performance of CDFlow against a range of widely-used optimization-based planners, all of which can effectively leverage the properties of the CDF. Our goal is to demonstrate that CDFlow enhances robustness and efficiency in complex scenarios where the original CDF and other baselines often struggle.

\paragraph{Setup}
We use motion planning datasets from MotionBenchMaker~\cite{chamzas2022-motion-bench-maker} for the 7-DoF Franka robot. These datasets contain a wide variety of complex scenes and thus provide a challenging benchmark for evaluating our approach. They consist of 10 different tasks, each with 100 problems, covering diverse planning scenarios such as bookshelves, kitchens, tables. Each problem specifies a collision-free initial joint configuration and defines the goal as a target joint configuration. For optimization-based comparisons, we integrate CDFlow into a gradient-based planner and benchmark it against representative solvers, including IPOP, QRQP, OSQP, and qpOASES. To represent obstacles, we uniformly sample points on the mesh surfaces in proportion to their surface area, with a density of 300 points per unit area, enabling controlled testing under different perception conditions by varying the point density.

\paragraph{Results}
The results in Table~\ref{tab:MP_bench} show that our CDFlow-based planner consistently outperforms the baselines across all ten tasks. On average, CDFlow achieves a success rate of 86\%, whereas the original CDF and other optimization-based solvers often fail in cluttered environments. In addition, CDFlow requires fewer optimization steps and lower joint-space goal error, demonstrating that improved robustness does not come at the expense of trajectory quality. Overall, the results confirm that CDFlow provides substantial advantages over existing distance field approaches in terms of success rate, convergence efficiency, and accuracy.

\section{Conclusion}
\label{sec:conclusion}

In this paper, we introduced CDFlow, a novel framework that resolves the gradient ambiguity and geometric distortion inherent in traditional Configuration Space Distance Fields (CDFs). By shifting the paradigm from a single-point estimate to modeling the entire distribution of minimal-distance collision configurations, our approach provides a robust, ambiguity-free gradient field for motion planning. We proposed a conditional Neural ODE-based architecture to learn this complex distribution, coupled with an adaptive sampling strategy to generate high-fidelity training data. 

Our extensive experiments demonstrated that CDFlow significantly outperforms baseline methods in challenging whole-body planning tasks. The results showed substantial improvements in success rate and planning efficiency, particularly in scenarios rich with multi-modal solutions and kinematic singularities. This work highlights the critical advantage of adopting a distributional perspective on configuration space geometry.

\paragraph{Limitations and Future Work.}
The current implementation of CDFlow relies on an offline data generation phase for each new scene, which can be time-consuming. Furthermore, the gradient computation during inference depends on Monte Carlo sampling, and its accuracy scales with the number of samples. Future work will focus on developing methods for online, incremental updates to the collision distribution, enabling adaptation to dynamic environments. We also plan to investigate more efficient, deterministic sampling strategies, such as quasi-Monte Carlo methods or learned proposal networks, to reduce the number of required samples for high-quality gradient approximation.

\bibliography{reference}

\begin{thebibliography}{10}

\bibitem{lavalle2006planning}
Steven~M LaValle.
\newblock {\em Planning algorithms}.
\newblock Cambridge university press, 2006.

\bibitem{lynch2017modern}
Kevin~M Lynch and Frank~C Park.
\newblock {\em Modern robotics}.
\newblock Cambridge University Press, 2017.

\bibitem{park2019deepsdf}
Jeong~Joon Park, Peter Florence, Julian Straub, Richard Newcombe, and Steven Lovegrove.
\newblock Deepsdf: Learning continuous signed distance functions for shape representation.
\newblock In {\em Proc.\ {IEEE} Conf.\ on Computer Vision and Pattern Recognition ({CVPR})}, pages 165--174, 2019.

\bibitem{ratliff2009chomp}
Nathan Ratliff, Matt Zucker, J~Andrew Bagnell, and Siddhartha Srinivasa.
\newblock {CHOMP}: Gradient optimization techniques for efficient motion planning.
\newblock In {\em Proc.\ {IEEE} Intl Conf.\ on Robotics and Automation ({ICRA})}, pages 489--494. IEEE, 2009.

\bibitem{schulman2014motion}
John Schulman, Yan Duan, Jonathan Ho, Alex Lee, Ibrahim Awwal, Henry Bradlow, Jia Pan, Sachin Patil, Ken Goldberg, and Pieter Abbeel.
\newblock Motion planning with sequential convex optimization and convex collision checking.
\newblock {\em The International Journal of Robotics Research}, 33(9):1251--1270, 2014.

\bibitem{driess2022learning}
Danny Driess, Jung-Su Ha, Marc Toussaint, and Russ Tedrake.
\newblock Learning models as functionals of signed-distance fields for manipulation planning.
\newblock In {\em Proc.\ Conference on Robot Learning ({CoRL})}, pages 245--255. PMLR, 2022.

\bibitem{li2024configuration}
Yiming Li, Xuemin Chi, Amirreza Razmjoo, and Sylvain Calinon.
\newblock Configuration space distance fields for manipulation planning.
\newblock {\em arXiv preprint arXiv:2406.01137}, 2024.

\bibitem{chen2018neural}
Ricky T.~Q. Chen, Yulia Rubanova, Jesse Bettencourt, and David Duvenaud.
\newblock Neural ordinary differential equations.
\newblock In {\em Advances in neural information processing systems}, volume~31, 2018.

\bibitem{grathwohl2018ffjord}
Will Grathwohl, Ricky~TQ Chen, Jesse Bettencourt, Ilya Sutskever, and David Duvenaud.
\newblock {FFJORD}: Free-form continuous dynamics for scalable reversible generative models.
\newblock In {\em International Conference on Learning Representations}, 2018.

\bibitem{vicini2022differentiable}
Delio Vicini, S{\'e}bastien Speierer, and Wenzel Jakob.
\newblock Differentiable signed distance function rendering.
\newblock {\em ACM Transactions on Graphics (TOG)}, 41(4):1--18, 2022.

\bibitem{mildenhall2021nerf}
Ben Mildenhall, Pratul~P Srinivasan, Matthew Tancik, Jonathan~T Barron, Ravi Ramamoorthi, and Ren Ng.
\newblock {NeRF}: Representing scenes as neural radiance fields for view synthesis.
\newblock {\em Communications of the ACM}, 65(1):99--106, 2021.

\bibitem{zucker2013chomp}
Matt Zucker, Nathan Ratliff, Anca~D Dragan, Mihail Pivtoraiko, Matthew Klingensmith, Christopher~M Dellin, J~Andrew Bagnell, and Siddhartha~S Srinivasa.
\newblock {CHOMP}: Covariant hamiltonian optimization for motion planning.
\newblock {\em The International Journal of Robotics Research}, 32(9-10):1164--1193, 2013.

\bibitem{izadi2011kinectfusion}
Shahram Izadi, David Kim, Otmar Hilliges, David Molyneaux, Richard Newcombe, Pushmeet Kohli, Jamie Shotton, Steve Hodges, Dustin Freeman, Andrew Davison, et~al.
\newblock {KinectFusion}: real-time 3d reconstruction and interaction using a moving depth camera.
\newblock In {\em Proceedings of the 24th annual ACM symposium on User interface software and technology}, pages 559--568, 2011.

\bibitem{weng2023ngdf}
Thomas Weng, David Held, Franziska Meier, and Mustafa Mukadam.
\newblock Neural grasp distance fields for robot manipulation.
\newblock {\em IEEE International Conference on Robotics and Automation (ICRA)}, 2023.

\bibitem{koptev2022neural}
Mikhail Koptev, Nadia Figueroa, and Aude Billard.
\newblock Neural joint space implicit signed distance functions for reactive robot manipulator control.
\newblock {\em IEEE Robotics and Automation Letters}, 8(2):480--487, 2022.

\bibitem{michaux2023reachability}
Jonathan Michaux, Qingyi Chen, Yongseok Kwon, and Ram Vasudevan.
\newblock Reachability-based trajectory design with neural implicit safety constraints.
\newblock {\em arXiv preprint arXiv:2302.07352}, 2023.

\bibitem{li2024representing}
Yiming Li, Yan Zhang, Amirreza Razmjoo, and Sylvain Calinon.
\newblock Representing robot geometry as distance fields: Applications to whole-body manipulation.
\newblock In {\em Proc.\ {IEEE} Intl Conf.\ on Robotics and Automation ({ICRA})}, pages 15351--15357, 2024.

\bibitem{Siciliano2008robotics}
Bruno Siciliano, Lorenzo Sciavicco, Luigi Villani, and Giuseppe Oriolo.
\newblock {\em Robotics: Modelling, Planning and Control}.
\newblock Springer Publishing Company, Incorporated, 1st edition, 2008.

\bibitem{ratliff2018riemannian}
Nathan~D Ratliff, Jan Issac, Daniel Kappler, Stan Birchfield, and Dieter Fox.
\newblock Riemannian motion policies.
\newblock {\em arXiv preprint arXiv:1801.02854}, 2018.

\bibitem{werner2023approximating}
Peter Werner, Alexandre Amice, Tobia Marcucci, Daniela Rus, and Russ Tedrake.
\newblock Approximating robot configuration spaces with few convex sets using clique covers of visibility graphs.
\newblock {\em arXiv preprint arXiv:2310.02875}, 2023.

\bibitem{lavalle2001randomized}
Steven~M LaValle and James~J Kuffner, Jr.
\newblock Randomized kinodynamic planning.
\newblock {\em The International Journal of Robotics Research}, 20(5):378--400, 2001.

\bibitem{williams2017model}
Grady Williams, Andrew Aldrich, and Evangelos~A Theodorou.
\newblock Model predictive path integral control: From theory to parallel computation.
\newblock {\em Journal of Guidance, Control, and Dynamics}, 40(2):344--357, 2017.

\bibitem{bhardwaj2022storm}
Mohak Bhardwaj, Balakumar Sundaralingam, Arsalan Mousavian, Nathan~D Ratliff, Dieter Fox, Fabio Ramos, and Byron Boots.
\newblock {STORM}: An integrated framework for fast joint-space model-predictive control for reactive manipulation.
\newblock In {\em Conference on Robot Learning}, pages 750--759. PMLR, 2022.

\bibitem{van2022geometric}
Karl Van~Wyk, Mandy Xie, Anqi Li, Muhammad~Asif Rana, Buck Babich, Bryan Peele, Qian Wan, Iretiayo Akinola, Balakumar Sundaralingam, Dieter Fox, et~al.
\newblock Geometric fabrics: Generalizing classical mechanics to capture the physics of behavior.
\newblock {\em IEEE Robotics and Automation Letters}, 7(2):3202--3209, 2022.

\bibitem{ratliff2015understanding}
Nathan Ratliff, Marc Toussaint, and Stefan Schaal.
\newblock Understanding the geometry of workspace obstacles in motion optimization.
\newblock In {\em 2015 IEEE International Conference on Robotics and Automation (ICRA)}, pages 4202--4209. IEEE, 2015.

\bibitem{klein2023design}
Holger Klein, No{\'e}mie Jaquier, Andre Meixner, and Tamim Asfour.
\newblock On the design of region-avoiding metrics for collision-safe motion generation on riemannian manifolds.
\newblock In {\em 2023 IEEE/RSJ International Conference on Intelligent Robots and Systems (IROS)}, pages 2346--2353. IEEE, 2023.

\bibitem{acar2021approximating}
Cihan Acar and Keng~Peng Tee.
\newblock Approximating constraint manifolds using generative models for sampling-based constrained motion planning.
\newblock In {\em 2021 IEEE International Conference on Robotics and Automation (ICRA)}, pages 8451--8457. IEEE, 2021.

\bibitem{ames2022ikflow}
Barrett Ames, Jeremy Morgan, and George Konidaris.
\newblock Ikflow: Generating diverse inverse kinematics solutions.
\newblock {\em IEEE Robotics and Automation Letters}, 7(3):7177--7184, 2022.

\bibitem{park2022node}
Suhan Park, Mathew Schwartz, and Jaeheung Park.
\newblock {NODE-IK}: Solving inverse kinematics with neural ordinary differential equations for path planning.
\newblock {\em arXiv preprint arXiv:2209.00498}, 2022.

\bibitem{mathieu2020riemannian}
Emile Mathieu, Tom Le~Lan, Chris~J. Maddison, Ryota Tomioka, and Yee~Whye Teh.
\newblock Riemannian continuous normalizing flows.
\newblock In {\em Advances in Neural Information Processing Systems (NeurIPS)}, volume~33, pages 20224--20235, 2020.

\bibitem{chamzas2022-motion-bench-maker}
Constantinos Chamzas, Carlos Quintero-Pe{\~n}a, Zachary Kingston, Andreas Orthey, Daniel Rakita, Michael Gleicher, Marc Toussaint, and Lydia E.~Kavraki.
\newblock Motionbenchmaker: A tool to generate and benchmark motion planning datasets.
\newblock {\em IEEE Robotics and Automation Letters}, 7(2):882–889, April 2022.

\end{thebibliography}

\end{document}